\documentclass[11pt,times]{article}
\usepackage{amsmath,amsfonts,amssymb,amsthm}
\usepackage{times}
\usepackage{geometry}
\usepackage{algorithm}
\usepackage{algorithmic}
\usepackage{amsmath}
\usepackage{amssymb}
\usepackage{graphicx}
\usepackage{subfigure}
\usepackage{hyperref}
\usepackage{url}
\usepackage{multirow}
\usepackage{amsmath,amsfonts,amssymb,amsthm}
\usepackage{algorithm}
\usepackage{algorithmic}
\usepackage{graphicx}
\usepackage{multirow}
\usepackage{wrapfig}
\usepackage{hyperref}

\makeatletter

\newtheorem{theorem}{Theorem}[section]
\newtheorem{lemma}[theorem]{Lemma}
\newtheorem{corollary}[theorem]{Corollary}

\theoremstyle{definition}

\theoremstyle{definition}

\newcommand{\squeezeup}{\vspace{-2.5mm}}

\makeatother

\begin{document}

\title{Online Convex Optimization Against Adversaries with  Memory and Application to Statistical Arbitrage}
\author{Oren Anava\\
\footnotesize{Technion, Haifa, Israel}\\ 
\footnotesize{oanava@tx.technion.ac.il}\\  \and Elad Hazan\\
\footnotesize{Technion, Haifa, Israel}\\ 
\footnotesize{ehazan@ie.technion.ac.il}  \and Shie Mannor\\
\footnotesize{Technion, Haifa, Israel}\\ 
\footnotesize{shie@ee.technion.ac.il}\\  }
\date{}
\maketitle

\begin{abstract}
The framework of online learning with memory naturally captures learning problems with temporal constraints, and was previously studied for the experts setting. In this work we extend the notion of learning with memory to the general Online Convex Optimization (OCO) framework, and present two algorithms that attain low regret.  The first algorithm applies to Lipschitz continuous loss functions, obtaining optimal regret bounds for both convex and  strongly convex losses. The second algorithm attains the optimal regret bounds and applies more broadly to  convex losses without requiring Lipschitz continuity, yet is more complicated to implement. 
We complement our theoretic results with an application to statistical arbitrage in finance: we devise algorithms for constructing mean-reverting portfolios.\end{abstract}

\section{Introduction}
\squeezeup
One of the most well-studied  frameworks of online learning is \emph{Online Convex Optimization} (OCO). In this framework, an online player iteratively chooses a decision in a convex set, then a convex loss function is revealed, and the  player suffers loss that is the convex function applied to the decision she  chose. It is usually assumed that the set of loss functions is chosen arbitrarily, possibly by an all-powerful adversary. The performance of the online player is  measured using the \emph{regret} criterion, which compares the accumulated  loss of the player with  the accumulated loss of  the best fixed decision in hindsight.

This notion of regret captures only \emph{memoryless adversaries} who determine the loss based on the player's current decision, and fails to cope with \emph{bounded-memory adversaries}  who determine the loss based on the player's current and previous decisions. However, in many scenarios such as coding, compression, portfolio selection and more, 
 the adversary is not completely memoryless and the previous decisions of the player affect her current loss. We are particularly concerned with scenarios in which the memory is relatively short-term and simple, in contrast to state-action models for which  reinforcement learning models  are more suitable \cite{puterman2009markov}. 
 
An important aspect of our work is that the memory is \emph{not} used to relax the adaptiveness of the adversary (cf.~\cite{DekelTA12,DBLP:journals/corr/abs-1302-4387}), but rather to model the feedback received by the player. In particular, throughout this work we assume a \textit{counterfactual feedback} model: the player is aware of the loss she would suffer had she played any sequence of  $m$ decisions in the previous $m$ time points. In addition, we assume that the adversary is \emph{oblivious}, that is, the adversary must determine the whole set of loss functions in advance. This model is quite common in the online learning literature \cite{DBLP:journals/tit/MerhavOSW02,MerhavOSW06,GyorgyN11}, yet was studied only for the experts problem.

Our goal in this work is to extend the notion of learning with memory to one of the most general online learning frameworks - the OCO. To this end, we adapt the \emph{policy regret}\footnote{The policy regret compares the performance of the online player with the best fixed sequence of actions in hindsight, and thus captures the notion of adversaries with memory. A formal definition appears in Section \ref{pre}.} criterion of \cite{DekelTA12}, and propose two different approaches for the extended framework, both attain the optimal bounds with respect to this criterion. 

We demonstrate the effectiveness of the proposed framework in the extensively studied problem of constructing mean-reverting portfolios.  
Specifically,  we cast this problem as an OCO problem with memory in which the loss functions are proxies for mean reversion, and the decisions of the player are wealth distributions over assets. The main novelty we present is the ability to maintain the wealth distributions online, in contrast to traditional approaches that determine the wealth distribution only at the end of the training period.  The experimental results support the superiority of our algorithm with respect to the state-of-the-art.

\subsection{Summary of Results}
\squeezeup

\begin{table}[h!] \label{state}
\begin{center}
\begin{tabular}{ |c|c|c|c| }
\hline
Framework & Previous bound & Our first approach & Our second approach \\ \hline
 {Experts} & \multirow{2}{*}{ $ \mathcal{O} ( T ^{1/2} ) $} & \multirow{2}{*}{ Not applicable } & \multirow{2}{*}{ $ \tilde{\mathcal{O}} ( T ^{1/2} ) $ } \\
{with Memory}  &  &  &  \\
 \hline
 OCO with memory& \multirow{2}{*}{ $  \mathcal{O} ( T ^{2/3} ) $} & \multirow{2}{*}{$ \mathcal{O} ( T ^{1/2} ) $} & \multirow{2}{*}{$ \tilde{\mathcal{O}} ( T ^{1/2} ) $} \\
(convex losses) &  &  &  \\
 \hline
  OCO with Memory & \multirow{2}{*}{ $  \tilde{\mathcal{O}} ( T ^{1/3} ) $} & \multirow{2}{*}{$ \mathcal{O} ( \log T ) $} & \multirow{2}{*}{$ \tilde{\mathcal{O}} ( T ^{1/2} ) $} \\
(strongly convex losses) &  &  &  \\
 \hline
\end{tabular}
\end{center}
\caption{State-of-the-art upper-bounds on the policy regret as a function of $T$ (number of iterations) for the framework of OCO with memory. The best known bounds are due to the works of \cite{GeulenVW10}, \cite{GyorgyN11}, and  \cite{DekelTA12}, which are detailed in the related work section below.}
\end{table}
We present and analyze two algorithms for the framework of OCO with memory, both attain {policy regret} bounds that are optimal in the number of iterations. 
Our first algorithm utilizes the Lipschitz property
of the loss functions, and --- to the best of our knowledge --- is the first algorithm for this framework that is not based on any blocking technique (this technique is detailed in the related work section below). This algorithm attains  $ \mathcal{O} (T^{1/2}) $-policy regret for generally convex loss functions and $ \mathcal{O} (\log T ) $-policy regret for strongly convex loss functions.
  
For the case of convex and non-Lipschitz loss functions, our  second algorithm attains the nearly optimal $\tilde{\mathcal{O}} ( T ^{1/2} ) $-policy regret;
its downside is that it is randomized and more difficult to implement.
A novel result that follows immediately from our analysis is that \textit{our second algorithm attains an expected $\tilde{\mathcal{O}} ( T ^{1/2} ) $-regret\footnote{The notation $\tilde{\mathcal{O}} ( \cdot ) $ is a variant of the $\mathcal{O}( \cdot ) $ notation that ignores logarithmic factors.}, along with $\tilde{\mathcal{O}} ( T ^{1/2} ) $ decision switches in the standard OCO framework}. Similar result currently exists only for the special case of the experts problem \cite{GeulenVW10}.

\section{Preliminaries and Model}  \label{pre}

We continue to formally define the notations for both the standard OCO framework and the framework of OCO with memory.
For sake of readability, we shall use the notations $g_t$ for memoryless loss functions (that correspond to memoryless adversaries), and $f_t$ for loss functions with memory (that correspond to bounded-memory adversaries). 

\subsection{The Standard OCO Framework}

In the standard OCO framework, an online player iteratively chooses a decision $ x_t \in \mathcal{K} $, and suffers loss that equals to $ g_t (x_t) $. The decision set $ \mathcal{K} $ is assumed to be a bounded convex subset of $ \mathbb{R}^n $, and the loss functions $ \{ g_t \} _{t=1}^T $ are assumed to be convex functions from $\mathcal{K}$ to $[0,1]$. In addition, the set $ \{ g_t \} _{t=1}^T $ is assumed to be chosen in advance, possibly by an all-powerful adversary that has full knowledge of our learning algorithm (see for instance \cite{CesaBianchiLu06}). The performance of the player is measured using the \emph{regret} criterion, defined as follows:
\begin{equation*} 
R_T = \sum_{t=1}^T g_t (x_{t})  - \min_{x \in \mathcal{K}} \sum_{t=1}^T g_t (x) ,
\end{equation*}
where $T$ is a predefined integer denoting the total number of iterations played. The goal in this framework is to design efficient algorithms, whose regret grows sublinearly in $T$, corresponding to an average per-round regret going to zero as $T$ increases. 


\subsection{The Framework of OCO with Memory}

In this work we consider the framework of OCO with memory, detailed as follows:
at each time point $t$, the online player chooses a decision $ x_t  \in \mathcal{K} \subset \mathbb{R}^n$. Then, a loss function $ f_t :  \mathcal{K} ^ {m+1} \rightarrow \mathbb{R} $ is revealed, and the player suffers loss that equals to $ f_t (x_{t-m}, \ldots , x_{t}) $. For simplicity of analysis we assume that $ 0 \in \mathcal{K}$, and that $ f_t (x_{0}, \ldots , x_{m}) \in [0,1] $ for any $ x_0,\ldots,x_m \in \mathcal{K} $. 
Notice that  the loss at time point $t$ depends on the previous $m$ decisions of the player, as well as on his current one. We assume that after $f_t$ is revealed, the player is aware of the loss she would suffer had she played any sequence of decisions $ x_{t-m}, \ldots, x_t $ (this correspond to the counterfactual feedback model mentioned earlier).

Our goal in this framework is to minimize the \emph{policy regret}, as defined in \cite{DekelTA12}\footnote{The iterations in which $t < m$ are ignored since we assume that the loss per iteration is bounded by a constant; this adds at most a constant to the final regret bound.}:
\begin{equation*}
R_{T,m} = \sum_{t=m}^T f_t (x_{t-m}, \ldots , x_{t})  - \min_{x \in \mathcal{K}} \sum_{t=m}^T f_t (x,\ldots,x) .
\end{equation*}
We define the notion of convexity for the loss functions $ \{ f_t \} _{t=1}^T $ as follows: we say that $f_t$ is a convex loss function with memory if $\tilde{f}_t (x)= f_t (x,\ldots,x)$ is convex in $x$. Throughout this work we assume that $ \{ f_t \} _{t=1}^T $ are convex loss functions with memory. This assumption can be shown to be necessary in some cases, if {\it efficient} algorithms are considered; otherwise, the optimization problem 
$
 \min_{x \in \mathcal{K}} \sum_{t=m}^T f_t (x,\ldots,x)
 $
 might be unsolvable efficiently.

\section{Policy Regret for  Lipschitz Continuous Loss Functions} \label{rftl_memory}
\squeezeup
In this section we assume that the loss functions $ \{ f_t \} _{t=1}^T $ are Lipschitz continuous for some Lipschitz constant $L$, that is $$ \left| f_t (x_0,\ldots,x_m)  -  f_t (y_0, \ldots , y_m) \right|  \leq  L \cdot \| (x_0,\ldots,x_m) - (y_0, \ldots , y_m) \| ,
$$
and adapt the well-known Regularized Follow The Leader (RFTL) algorithm to cope with bounded-memory adversaries.  We present here only the algorithm and the main theorem, and defer the complete analysis to Appendix \ref{rftl_app}.

\begin{algorithm}[H]
\caption{RFTL with Memory (RFTL-M)}
\label{alg:rftlm}
\begin{algorithmic}[1]
\STATE Input: learning rate $\eta$, regularization function $ \mathcal{R} (x)$, loss functions with memory $\{ f_t \}_{t=1}^T$.
\STATE Choose $x_0, \ldots , x_m \in \mathcal{K}$ arbitrarily.
\FOR {$t=m$ to $T$}
\STATE Play $x_t$ and suffer loss $ f_t (x_{t-m}, \ldots , x_{t}) $.
\STATE Set  $x_{t+1} = \arg \min_{x \in \mathcal{K}} \left\{ \eta \cdot \sum_{\tau = 1}^{t} \tilde{f}_{\tau} (x) + \mathcal{R} (x) \right\} $
\ENDFOR
\end{algorithmic}
\end{algorithm}

Intuitively, Algorithm  \ref{alg:rftlm} relies on the fact that the corresponding  functions $ \{ \tilde{f}_t \}_{t=1}^T $ are memoryless and convex. Thus, standard regret minimization techniques are applicable, yielding a regret bound of $\mathcal{O} ( T^{1/2} ) $ for $ \{ \tilde{f}_t \}_{t=1}^T $. This however, is not the policy regret bound we are interested in, but is in fact quite close if we use the Lipschitz property of $ \{ {f}_t \}_{t=1}^T $ and set the learning parameter properly. 
 For Algorithm \ref{alg:rftlm} we can prove the following:\\
\begin{theorem} \label{rftlm}
Let $\{ f_t \}_{t=1}^T$ be Lipschitz continuous loss functions with memory (from $\mathcal{K}^{m+1}$ to $[0,1]$), and let $R$ and $\lambda$ be  as defined in Equation \eqref{params}. Then, Algorithm \ref{alg:rftlm} generates an online sequence $\{ x_t \} _{t=1}^T $, for which the following holds:
\begin{equation*}
R_{T,m} = \sum_{t=m}^T f_t (x_{t-m}, \ldots , x_{t})   -    \min_{x \in \mathcal{K}} \sum_{t=m}^T f_t (x,\ldots,x) \leq 4 T \lambda \eta m^{3/2}  + \frac{R}{\eta} .
\end{equation*}
Setting $ \eta = \sqrt{\frac{R}{4 T \lambda m^{3/2}}} $ yields $ R_{T,m} \leq 4 \sqrt{TR \lambda  m^{3/2}}$ .
\end{theorem}


\squeezeup
\section{Policy Regret with Low Switches } \label{low}
\squeezeup
In this section we present a different approach to the framework of OCO with memory --- low switches. This approach was considered before in \cite{GyorgyN11}, who adapted the Shrinking Dartboard (SD) algorithm of \cite{GeulenVW10} to cope with limited-delay coding. However, in \cite{GeulenVW10,GyorgyN11} consider only the experts setting, in which the decision set is the simplex and the loss functions are linear. Here we adapt this approach to general decision sets and generally convex loss functions, and obtain optimal policy regret against bounded-memory adversaries. 
We present here only the algorithm and main theorem, and defer the complete analysis to Appendix \ref{low_app}.

\begin{algorithm}[H]
\caption{}
\label{alg:rewoo_low}
\begin{algorithmic}[1]
\STATE Input: learning parameter $\eta$, convex loss functions $\{ g_t \} _{t=1}^T $.
\STATE Initialize $w_1 (x) =1$ for all $x\in\mathcal{K}$, and choose $x_1 \in \mathcal{K}$ arbitrarily.
\FOR {$t=1$ to $T$}
\STATE Play $x_t$ and suffer loss $ g_t (x_{t}) $.
\STATE Define weights $w_{t+1} (x)= e^{- \alpha \sum_{\tau=1}^{t} \hat{g}_{\tau} (x)} $, where $\alpha=\frac{\eta}{4G^2}$ and $\hat{g}_t(x) = g_t (x) + \frac{\eta}{2} \| x \|^2$.
\STATE Set $x_{t+1} = x_t$ with probability $ \frac{w_{t+1}(x_t)}{w_t(x_t)}$.
\STATE Otherwise, sample $x_{t+1} $ from the density function $ p_{t+1}(x) = w_{t+1} (x) \cdot \left(  \int_{\mathcal{K}} w_{t+1} (x) dx  \right)^{-1}$.
\ENDFOR
\end{algorithmic}
\end{algorithm}
Intuitively, Algorithm \ref{alg:rewoo_low} defines a probability distribution over $\mathcal{K}$ at each time point $t$. By sampling from this probability distribution one can generate an online sequence that has an expected  low regret guarantee. This however is not sufficient in order to cope with bounded-memory adversaries, and thus an additional element of choosing $x_{t+1} = x_t$ with high probability is necessary (line 6). Our analysis shows that if this probability equals to $ \frac{w_{t+1}(x_t)}{w_t(x_t)}$ the regret guarantee remains, and we get an additional low switches guarantee.\\
   
 For Algorithm \ref{alg:rewoo_low} we can prove the following:
\begin{theorem} \label{main2}
Let $\{ g_t \}_{t=1}^T$ be convex functions from $\mathcal{K} $ to $ [0,1] $, such that 
$D =  \sup_{x,y \in \mathcal{K}} \| x-y\| $ and $ G =  \sup_{x,t} \| \nabla  g_t (x)  \| $,
 and define $\hat{g}_t(x) = g_t(x) + \frac{\eta}{2} \| x\|^2$ for some $\eta \leq \frac{G}{D}$.  Then,  Algorithm \ref{alg:rewoo_low}  generates an online sequence $\{ x_t\}_{t=1}^T$, for which it holds that
$$ \mathbb{E} \left[ R_T \right] = \sum_{t=1}^T \mathbb{E} \left[ g_t (x_{t})  \right]  - \min_{x \in \mathcal{K}} \sum_{t=1}^T g_t (x)  \leq \frac{4G^2}{\eta}  \left( 1 + n \log(T+1) \right) + \frac{T \eta }{2}  \left( \frac{ \left( 1+\eta D^2  \right)^2 }{4G^2} +  D^2   \right) ,$$
and in addition
$$ \mathbb{E} \left[ S \right]  =\mathbb{E} \left[  \sum_{t=1}^T 1_{ \{ x_{t+1} \neq x_{t} \} }  \right]  \leq \frac{T \eta}{4G^2} + \frac{T D^2 \eta^2}{8G^2}  ,  $$
where $S$ denotes the number of decision switches in the sequence $\{ x_t\}_{t=1}^T$.

\noindent Setting $\eta = \frac{2G}{D} \sqrt{ \frac{1+ \log(T+1)}{T} } $ yields $\mathbb{E} \left[ R_T \right]  = \mathcal{O} \big( \sqrt{ T \log(T) } \big)$, and $ \mathbb{E} \left[ S \right] = \mathcal{O} \big( \sqrt{ T \log(T) } \big)$.
\end{theorem}

Notice that Algorithm \ref{alg:rewoo_low} applies to memoryless loss functions, yet its low switches guarantee implies learning against bounded-memory adversaries as stated and proven in Lemma \ref{reduction} (Appendix \ref{reduce}).

\squeezeup
\section{Application to Statistical Arbitrage} \label{app}
\squeezeup
Our application is motivated by financial models that are aimed at creating statistical arbitrage opportunities. 
In the literature, ``statistical arbitrage'' refers to statistical mispricing of one or more assets based on their expected value.  One of the most common trading strategies, known as ``pairs trading'', seeks to create a mean reverting portfolio using two assets with same sectoral belonging (typically using both long  and short sales). Then, by buying this portfolio below its mean and selling it above, one can have an expected positive profit with low risk. 

Here we extend the traditional pairs trading strategy, and present an approach that aims at constructing a mean reverting portfolio from an arbitrary (yet known in advance) number of assets. Roughly speaking, our goal is to synthetically create a mean reverting portfolio by maintaining weights upon $n$ different assets. The main problem arises in this context is how do we quantify the amount of mean reversion of a given portfolio? Indeed, mean reversion is somewhat an ill-defined concept, and thus different proxies are usually defined to capture its notion. We refer the reader to \cite{Schmidt2011,daspremont2011}, in which few of these proxies (such as predictability and zero-crossing) are presented.

In this work, we consider a proxy that is aimed at preserving the mean price of the constructed portfolio (over the last $m$ trading periods) close to zero, while maximizing its variance. 
We note that due to the very nature of the problem: weights of one trading period affect future performance, the memory comes unavoidably into the picture. 

 We proceed to formally define the new mean reversion proxy and the use of our new memory-learning algorithm in this model.
 Denote by $ y_t \in \mathbb{R}^n $ the prices of $n$ assets at time $t$, and by $ x_t \in \mathbb{R}^n $ a distribution of  weights over these assets. Since short selling is allowed, the norm of $x_t$ can sum up to an arbitrary number, determined by the loan flexibility. Without loss of generality we assume that $ \| x_t \| _2 = 1 $, and define:
\begin{equation} \label{proxy}
f_t (x_{t-m} , \ldots , x_t )   = \left( \sum_{i=0}^{m} x_{t-i}^\top y_{t-i} \right)^2 - \lambda \cdot  \sum_{i=0}^{m} \left(x_{t-i}^\top y_{t-i} \right) ^2  ,
\end{equation}
for some $\lambda>0$. Notice that minimizing $f_t$ iteratively  yields a process $ \{ x_t^\top y_t \} _{t=1}^T $ such that its mean is close to zero (due to the expression on the left), and its variance is maximized (due to the expression on the right). 
We use the regret criterion to measure our performance against the best distribution of weights in hindsight, and wish to generate a series of weights $\{x_t\}_{t=1}^T$ such that the regret is sublinear.
Thus, define the memoryless loss function $\tilde{f}_t (x) = f_t (x,\ldots,x)$ and denote
$$ A_t = \sum_{i=0}^{m-1}  \sum_{j=0}^{m-1}  y_{t-i}   y_{t-j} \quad \text{ and } \quad B_t = \lambda \cdot \left( \sum_{i=0}^{m-1} y_{t-i} y_{t-i}^T \right). $$
Notice we can write $ \tilde{f}_t (x) = x^\top A_t x - x^\top B_t x $. Since $\tilde{f}_t$ is not convex in general, our techniques are not straightforwardly applicable here. However, the hidden convexity of the problem allows us to bypass this issue by a simple and tight Positive Semi-Definite (PSD) relaxation. Define
\begin{equation} \label{ht}
h_t (X) =X \circ A_t - X \circ B_t ,
\end{equation}
where $X$ is a PSD matrix with $Tr(X) = 1$, and $ X \circ A $ is defined as $ \sum_{i=1}^n  \sum_{j=1}^n X(i,j) \cdot A(i,j) $.
Now, notice that the problem of minimizing $\sum_{t=m}^T h_t (X)$ is a PSD relaxation to the minimization problem  $\sum_{t=m}^T \tilde{f}_t (x)$, and for the optimal solution it holds that:
\[
\min_{X} \sum_{t=m}^T h_t (X)    \leq   \sum_{t=m}^T h_t (x^* x^{* \top})    =   \sum_{t=m}^T \tilde{f}_t (x^*) .
\]
where $x^* = \arg \min_{x \in \mathcal{K} } \sum_{t=m}^T \tilde{f}_t (x) $. Also, we can recover a vector $x$ from the PSD matrix $X$ using an eigenvector decomposition as follows: represent $X = \sum_{i=1}^n \lambda_i v_i v_i^\top$, where each $v_i$ is a unit vector and  $\lambda_i$ are non-negative coefficients such that $\sum_{i=1}^n \lambda_i =1$. Then, by sampling the eigenvector $x = v_i$ with probability $\lambda_i$, we get that $ \mathbb{E} \big[ \tilde{f}_t (x) \big]  =  h_t (X) $. 
Technically, this decomposition is possible due to the fact that $X$ is a PSD matrix with $Tr(X) = 1$. 
Notice that $h_t$ is linear in $X$, and thus we can apply regret minimization techniques on the loss functions $\{ h_t \} _{t=1}^T $. 
This procedure is formally given in Algorithm \ref{alg:osa} below.

\begin{algorithm}[H]
\caption{Online Statistical Arbitrage (OSA)}
\label{alg:osa}
\begin{algorithmic}[1]
\STATE Input: Learning rate $\eta$, memory parameter $m$, regularizer $\lambda$.
\STATE Initialize $X_1 = \frac{1}{n} I_{n \times n} $.
\FOR {$t=1$ to $T$}
\STATE Randomize $x_{t} \sim X_t $ using the eigenvector decomposition.
\STATE Observe $f_t$ and define $h_t$ as in equation \eqref{ht}.
\STATE Apply Algorithm \ref{alg:rewoo_low}  to $h_t (X_t)$ to get   $X_{t+1}$.
\ENDFOR
\end{algorithmic}
\end{algorithm}

For Algorithm \ref{alg:osa} we can prove the following:

\begin{corollary} 
Let $\{ f_t \}_{t=1}^T$ be as defined in Equation \eqref{proxy}, and $\{ h_t \} _{t=1}^T$ be the corresponding memoryless functions, as defined in Equation \eqref{ht}. 
Then, applying Algorithm \ref{alg:rewoo_low} to the loss functions  $\{ h_t \}_{t=1}^T$  yields an online sequence $\{ X_t\}_{t=1}^T$, for which the following holds:
$$
 \sum_{t=1}^T \mathbb{E} \left[  h_t (X_t) \right] - \min_{ \substack{ X \succeq 0 \\\ \text{Tr}(X)=1} }   \sum_{t=1}^T h_t (X)  =  \mathcal{O} \big( \sqrt{ T \log(T) } \big) .
$$
Sampling $x_t \sim X_t$ using the eigenvector decomposition described above yields:
$$
 \mathbb{E} \left[ R_{T,m} \right] =  \sum_{t=m}^T  \mathbb{E} \left[ f_t (x_{t-m},\ldots,x_t) \right] - \min_{ \| x \|=1}   \sum_{t=m}^T f_t (x,\ldots,x)  =  \mathcal{O} \big( \sqrt{ T \log(T) } \big) .
$$
\end{corollary}

The main novelty of our approach to the task of constructing mean reverting portfolios is the ability to maintain the weight distributions online. This is in contrast to the traditional offline approaches that require a training period (to learn a weight distribution), and a trading period (to apply a corresponding trading strategy). 
\squeezeup
\section{Experimental Results}
\squeezeup
In this section we present some preliminary results that demonstrate the effectiveness of the proposed algorithm to the task of creating statistical arbitrage opportunities under the pairs trading setting. In this setting, we are given two assets with the same sectoral belonging and our goal is to construct a mean reverting portfolio by maintaining weights upon these assets. To simplify the setting we ignore transaction costs (both for our algorithm and the benchmarks).

In order to isolate the problem of constructing a mean reverting portfolio (which is of our interest) from the problem of designing a trading strategy, the experiments are executed in two stages: first, a mean reverting portfolio is constructed by each of the considered approaches (which are described below in Section \ref{baseline}). Then, the same trading strategy is applied to all resulted portfolios, so that the different approaches are comparable in terms of return.  

Our dataset contains time series of daily closing rates of 10 pairs of assets based on their common sectoral belonging (e.g., Coca Cola and Pepsi, AT\&T and Verizon, etc.). We use data between 01/01/2008 and 01/02/2013, which is divided into training set (75\% of the data, from 01/01/2008 to 01/10/2011) and test set (25\% of the data, from 02/10/2011 to 01/02/2013). 

\subsection{Baselines} \label{baseline}
\squeezeup
 In order to capture the essence of our Online Statistical Arbitrage (OSA) algorithm with respect to its offline counterparts, we choose some of the fundamental offline approaches\footnote{We refer the reader to \cite{maddala1998unit,Johansen91} for more comprehensive information about \textbf{OLS} and \textbf{Johansen}.} to serve as benchmarks:
\begin{description}
 \item \textbf{Orthogonal Least Squares (OLS)}  this baseline proposes to choose the eigenvector that corresponds to smallest eigenvalue of the empirical covariance matrix of $y_t$. This matrix is denoted by $C$, and formally defined as follows:
$$
C =  \frac{1}{T_{\text{training}}-1} \cdot \sum_{t=1}^{T_{\text{training}}} \tilde{y}_t \tilde{y}_t^\top  \ \quad , \ \quad \text{where} \ \quad \tilde{y}_t = y_t - \frac{1}{T_{\text{training}}} \cdot \sum_{t=1}^{T_{\text{training}}} {y}_t ,
$$
 where $T_{\text{training}}$ denotes the number of days in the training set.
\item \textbf{Johansen Vector Error Correction Model}  this baseline relies on co-integration techniques. Basically, co-integration is a statistical relationship where two time series (e.g., stock prices) that are both integrated of same order $d$ can be linearly combined to produce a single time series which is integrated of order $d-b$, where $b>0$. In its application to pairs trading, the co-integration technique seeks to find a linear combination such that $d=b=1$, which roughly results in a mean reverting combined asset. 
\item \textbf{The offline optimum (Offline)} this baselines refers to the best distribution of weights in hindsight with respect to our proxy, that is
\[
x_{\text{Offline}} = \arg\min_x \left\{ \sum_{t=m}^{T_{\text{test}}} \left( \sum_{i=0}^{m} x^T y_{t-i} \right)^2 - \lambda \cdot \sum_{i=0}^{m} \left(x^T y_{t-i} \right) ^2 \right\} .
 \]
Here, $T_{\text{test}}$ denotes the number of days in the test set.
 Clearly, the performance of this baselines cannot be obtained in practice, as it relies on the future prices of the considered assets when constructing the portfolio. Nevertheless, this baseline has a crucial role in understanding the effectiveness of the proposed mean reversion proxy.
 \end{description}
 For the \textbf{OLS} and \textbf{Johansen} baselines we use the training period to generate a weight distribution $x$, and then construct the portfolio $\{ x^\top y_t \}_{t=1}^{T_\text{test}}$. For \textbf{OSA} we run Algorithm \ref{alg:osa} on the training set to get the sequence $\{ x_t \}_{t=1}^{T_\text{training}}$. Then, we use $x_{T_\text{training}}$ as a warm start for a new run of Algorithm \ref{alg:osa} on the test data to generate the portfolio $\{ x_t^\top y_t \}_{t=1}^{T_\text{test}}$ (which will be used for the benchmark task). 

\subsection{Trading Strategy}

\squeezeup

\begin{figure}[t]
     \begin{center}
     \squeezeup
           \includegraphics[width=\columnwidth]{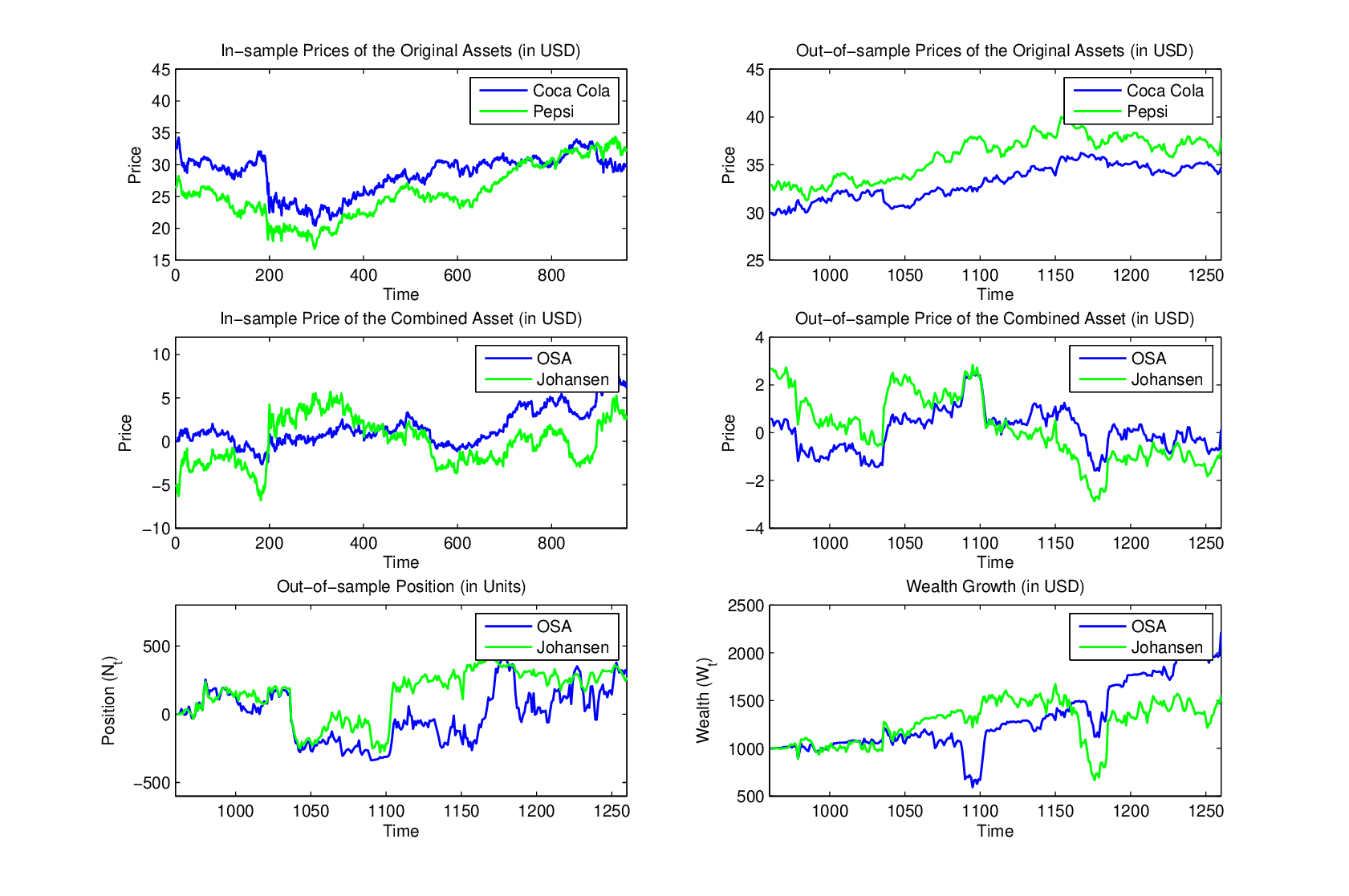}
           \squeezeup
    \caption{\textbf{Sample experimental results of OSA and Johansen for the pair Coca Cola and Pepsi}}
    \squeezeup
               \label{fig:1}
         \end{center}
         \squeezeup
\end{figure}

In order to compare the different approaches, we apply the trading strategy of \cite{jurek2007dynamic} to each of the resulting portfolios. Basically, \cite{jurek2007dynamic} propose to take a position $N_t$ in the asset $z_t$ proportionally to $\frac{\alpha (\mu - z_t) }{\sigma^2} W_t$, where $W_t$ denotes the wealth at time $t$ and  $\{z_t\}_{t=1}^T$ is assumed to be an auto regressive process of order 1 with mean $\mu$ that complies with
$ z_{t+1} = \alpha z_t + \sigma \epsilon_t $
(and  $\epsilon_t \sim \mathcal{N}(0,1)$). Essentially, this strategy takes a long position whenever the asset is below its mean and short position whenever it is above, while taking into account the autoregressive model parameters $\alpha$ and $\sigma$. 
In practice, these parameters are estimated on the training set and then used to generate $N_t$. A sample experiment  for the pair Coca Cola and Pepsi (using the entire training and test sets) that compares the performance of our algorithm and Johansen's is illustrated in Figure \ref{fig:1}.

\subsection{Results}
\begin{figure} [H]
\squeezeup
  \begin{minipage}[b]{0.32\linewidth}
    \begin{tabular}{c|c|c|} 
 \cline{2-3} 
& \multicolumn{2}{ |c| }{\small{Return (in \%)}} \\ \cline{2-3} 
 & \scriptsize{8-month} & \scriptsize{16-month}\\ \cline{1-3} 
\multicolumn{1}{ |c| }{\small{\textbf{Offline}}} &     \small{  \textbf{39.45} }&    \small{   \textbf{102.67}} \\ \cline{1-3} 
\multicolumn{1}{ |c| }{\small{\textbf{OSA}}} &       \small{ 33.59 }&     \small{   98.33 }\\ \cline{1-3} 
\multicolumn{1}{ |c| }{\small{\textbf{OLS}}} &       \small{ 23.64 }&      \small{  83.68 }\\ \cline{1-3} 
\multicolumn{1}{ |c| }{\small{\textbf{Johansen}}} &  \small{   33.87} &      \small{ 60.47} \\ \cline{1-3} 
\end{tabular}

    \par\vspace{14.5pt}
  \end{minipage}
    \begin{minipage}[b]{0.71\linewidth}
    \includegraphics[width=\linewidth]{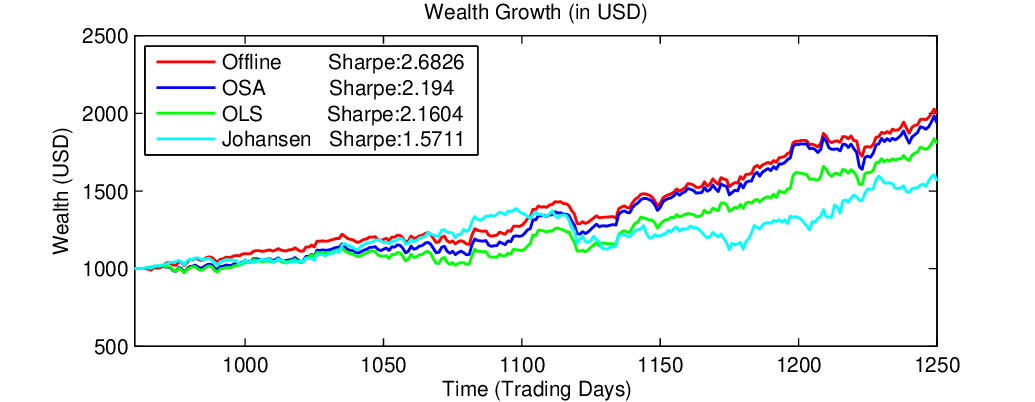}
    \par\vspace{0pt}
  \end{minipage}%
\caption{\textbf{Wealth as a function of time for the online algorithm and the three offline baselines}}
\label{fig:2}
\end{figure}

\squeezeup
In Figure \ref{fig:2} we plot the cumulative wealth of our online algorithm and the three offline baselines, and also provide the Sharpe ratios. To execute this experiments we use the 10 pairs of assets in our dataset. In all runs of our online algorithm and its offline counterpart we set $m=5$ and $\lambda=1$, arbitrarily. The task of determining the best values of $m$ and $\lambda$ is outside the scope of this paper, yet is a very challenging problem. The empirical observations clearly verify the effectiveness of the proposed mean reversion proxy and the online algorithm, as both \textbf{OSA} and \textbf{Offline} outperform the other baselines.
It can can also be seen that the performance of \textbf{OSA} approaches the performance of \textbf{Offline} as time advances, corresponding to our theoretic regret guarantee. 
 It remains for future work to compare the performance of the online approach and the offline state-of-the-art approaches in the presence of transaction costs.

\bibliography{memoryOCO}
\bibliographystyle{alpha}

\appendix

\section{Complete Analysis for Section \ref{rftl_memory}} \label{rftl_app}
We start by providing some  necessary background, and then turn to state and prove our main theorem. We complement our analysis with the special case in which the loss functions are  strongly convex (Appendix \ref{ogd}).
\subsection{Background} \label{back}
Recall the RFTL algorithm, which is one of the most popular algorithms for the standard OCO framework. Basically, RFTL generates the decision at time point $t$ according to the following rule:
$$
x_t = \arg \min_{x \in \mathcal{K}} \left\{ \eta \cdot \sum_{\tau = 1}^{t-1} g_{\tau} (x) + \mathcal{R} (x) \right\},
$$
where $ \eta$ is a predefined learning parameter, and $ \mathcal{R} (x) $ is called a regularization function. Note that $ \mathcal{R} (x) $ is chosen by the online player, and assumed to be $\sigma$-strongly convex\footnote{ The function $ \mathcal{R} (x) $ is called $\sigma$-strongly convex if $\nabla^2 \mathcal{R} (x) \succeq \sigma \cdot I_{d \times d}$ for all $x \in \mathcal{K}$.} and smooth, such that its second derivative is continuous.

Usually, general matrix norms are used to analyze and bound the regret of the RFTL algorithm: a PSD matrix $A \succ 0$ gives rise
to the norm $\left\| x \right\| _{A} = \sqrt{x^\top A x}$; its dual norm is $ \left\| x \right\| _{A^{-1}} = \left\| x \right\| _{A}^{*}$.
In particular, the interesting case is when $ A = \nabla^2 \mathcal{R} $, the Hessian of the regularization function. In this case, the notation is shorthanded to be $\left\| x \right\| _{\nabla^2 \mathcal{R} (y) } = \left\| x \right\| _y$ and  $\left\| x \right\| _{\nabla^{-2} \mathcal{R} (y) } = \left\| x \right\| _y^*$.

Now, if we denote  $$ \lambda = \sup_{t \in \{ 1, \ldots , T\},x,y \in \mathcal{K}} \left\{ \left( \| \nabla g_t ( x ) \|_y ^* \right) ^2 \right\} \ \ \text{ and }  \ \ R = \sup_{ x,y \in \mathcal{K} } \left\{ \mathcal{R} (x) - \mathcal{R} (y) \right\} ,$$ 
then, the RFTL algorithm generates an online sequence $\{x_t\} _{t=1}^T$, for which the following holds:
\begin{equation} \label{rftl_regret}
R_T = \sum_{t=1}^T g_t ( x_t ) - \min_{x \in \mathcal{K}}  \sum_{t=1}^T g_t ( x ) \leq 2 T \lambda   \eta + \frac{R}{\eta} .
\end{equation}
A complete analysis can be found in \cite{hazan201110,Shalev-Shwartz12}.

 \subsection{Adapting RFTL to the Framework of OCO with Memory}
We start by defining the function $\tilde{f}_t$ as follows: $ \tilde{f}_t (x) = f_t (x,\ldots,x)$. Recall that $ \tilde{f}_t (x)$ is convex in $x$ for all $t$, as assumed in Section \ref{pre}.  Following the notations of Section \ref{back}, we define a regularization function $ \mathcal{R} (x) $ and upper-bound
\begin{equation} \label{params}
 \lambda = \sup_{t \in \{ 1, \ldots , T\},x,y \in \mathcal{K}} \left\{ \left( \| \nabla \tilde{f}_t ( x ) \|_y ^* \right) ^2 \right\} \ \ \text{ and }  \ \ R = \sup_{ x,y \in \mathcal{K} } \left\{ \mathcal{R} (x) - \mathcal{R} (y) \right\} .
 \end{equation}
Notice that $\lambda$ might depend implicitly on $m$.
It follows that the loss functions $ \{ \tilde{f}_t \} _{t=1}^T $ are Lipschitz continuous for the Lipschitz constant $ \sqrt{\lambda \sigma} $ with respect to the $ \ell_2 $-norm. I.e., it holds that
$$
\big|  \tilde{f}_t (x) - \tilde{f}_t (y)  \big|  \leq  \sqrt{ \lambda \sigma} \cdot \| x - y \| .
$$
Without loss of generality, we can assume that the loss functions $ \{ f_t \} _{t=1} ^T $ are Lipschitz continuous for the same constant, i.e.,
$$ \left| f_t (x_0,\ldots,x_m)  -  f_t (y_0, \ldots , y_m) \right|  \leq   \sqrt{\lambda \sigma} \cdot \| (x_0,\ldots,x_m) - (y_0, \ldots , y_m) \| .
$$
Otherwise, we can simply set 
 $\lambda$  to satisfy this condition.

 The following is our main theorem, stated and proven:
 
 \textbf{Theorem 3.1.} 
\textit{Let $\{ f_t \}_{t=1}^T$ be Lipschitz continuous loss functions with memory (from $\mathcal{K}^{m+1}$ to $[0,1]$), and let $R$ and $\lambda$ be  as defined in Equation \eqref{params}. Then, Algorithm \ref{alg:rftlm} generates an online sequence $\{ x_t \} _{t=1}^T $, for which the following holds:}
\begin{equation*}
R_{T,m} = \sum_{t=m}^T f_t (x_{t-m}, \ldots , x_{t})   -    \min_{x \in \mathcal{K}} \sum_{t=m}^T f_t (x,\ldots,x) \leq 4 T \lambda \eta m^{3/2}  + \frac{R}{\eta} .
\end{equation*}
\textit{Setting $ \eta = \sqrt{\frac{R}{4 T \lambda m^{3/2}}} $ yields $ R_{T,m} \leq 4 \sqrt{TR \lambda  m^{3/2}}$.}

\begin{proof}
First, note that applying Algorithm \ref{alg:rftlm} to the loss functions $\{ f_t\}_{t=1}^T$ is equivalent to applying the original RFTL algorithm to the loss functions $\{\tilde{f}_t\}_{t=1}^T$. I.e., given $m$ initial points $ x_1, \ldots, x_m $, both algorithms generate the same sequence of decisions $\{ x_t\}_{t=m}^T$, for which it holds that:
$$
\sum_{t=m}^T \tilde{f}_t (x_{t})   -    \min_{x\in \mathcal{K}} \sum_{t=m}^T \tilde{f}_t (x) \leq 2 T \lambda  \eta + \frac{R}{\eta} ,
$$
or equivalently:
\begin{equation} \label{reg_rftl}
\sum_{t=m}^T f_t (x_{t}, \ldots , x_{t})   -    \min_{x\in \mathcal{K}} \sum_{t=m}^T f_t (x,\ldots,x) \leq 2 T \lambda  \eta + \frac{R}{\eta} ,
\end{equation}
due to the regret guarantee in Equation \eqref{rftl_regret}.
On the other hand, ${f}_t$ is Lipschitz continuous for the Lipschitz constant $\sqrt{\lambda \sigma}$, and thus we can bound
\begin{eqnarray*}
\left| f_t (x_t,\ldots,x_t)  -  f_t (x_{t-m}, \ldots , x_{t}) \right| ^2 &\leq & \left( \sqrt{\lambda \sigma} \cdot \| (x_t,\ldots,x_t) - (x_{t-m}, \ldots , x_{t}) \| \right) ^2  \\
& = & \lambda \sigma \cdot \sum_{j=1}^{m} \| x_t - x_{t-j} \| ^2 \\
& \leq &  \lambda \sigma \cdot \sum_{j=1}^{m} \left( \sum_{l=1}^{j}  \| x_{t-l+1} - x_{t-l}\| \right) ^2 \\
& \leq & \lambda \sigma \cdot \sum_{j=1}^{m} \left( \sum_{l=1}^{j} \frac{1}{\sqrt{\sigma}} \| x_{t-l+1} - x_{t-l}\|_{z_{t-l}} \right) ^2 \\
& \leq & \lambda \sigma \cdot \sum_{j=1}^{m} \left( \sum_{l=1}^{j} \frac{2 \eta \sqrt{\lambda}}{\sqrt{\sigma}}  \right) ^2 \leq  \lambda \sigma \cdot \sum_{j=1}^{m} \left(  \frac{4 m^2 \eta^2 \lambda }{\sigma}  \right) \\
& \leq & 4 \lambda^2 \eta ^2 m^3  ,
\end{eqnarray*}
where $z_t \in \left[ x_t , x_{t+1} \right] $. The inequality $ \| x_{t+1} - x_{t}\|_{z_{t}} \leq 2 \eta \sqrt{\lambda}  $ follows from the standard analysis of the RFTL algorithm \cite{hazan201110}. 
It follows that 
$
 \left| f_t (x_t,\ldots,x_t)  -  f_t (x_{t-m}, \ldots , x_{t}) \right|  \leq 2 \lambda \eta m^{3/2} ,
$
and by summing over $t=m,\ldots,T$ we get that
\begin{equation} \label{lip_rftl}
\left| \sum_{t=m}^T f_t (x_{t}, \ldots , x_{t}) - \sum_{t=m}^T f_t (x_{t-m}, \ldots , x_{t}) \right|   \leq 2 T \lambda \eta m^{3/2}  .
\end{equation}
Next, by integrating Equations \eqref{reg_rftl} and \eqref{lip_rftl} and using the fact that $m\geq1$ in our setting, we have that
\begin{equation*}
R_{T,m} = \sum_{t=m}^T f_t (x_{t-m}, \ldots , x_{t})   -    \min_{x\in \mathcal{K}} \sum_{t=m}^T f_t (x,\ldots,x) \leq 4 T \lambda \eta m^{3/2} + \frac{R}{\eta} .
\end{equation*}
Finally, setting $ \eta = \sqrt{\frac{R}{4 T \lambda m^{3/2}}} $ yields
\begin{equation*}
R_{T,m} = \sum_{t=m}^T f_t (x_{t-m}, \ldots , x_{t})   -    \min_{x\in \mathcal{K}} \sum_{t=m}^T f_t (x,\ldots,x) \leq 4  \sqrt{T R \lambda m^{3/2}} ,
\end{equation*}
as stated in the theorem.
\end{proof}

\subsection{Extending Algorithm \ref{rftlm} to Strongly Convex Loss Functions } \label{ogd}
In the standard OCO framework, it is well known that plugging  $ \mathcal{R}(x) = \| x \| ^2 $ in the RFTL algorithm yields the familiar Online Gradient Descent (OGD) algorithm of \cite{Zinkevich03}. In this case, it is easy to show that $R=\mathcal{O}\left(D^2\right)$ and $\lambda=\mathcal{O}\left(G^2\right)$, where $ D= \sup_{x,y \in \mathcal{K}} \| x-y\|  $ and $ G= \sup_{t,x \in \mathcal{K}} \| \nabla  g_t (x)  \|  $ . Substituting these values in Equation \eqref{rftl_regret} results in the following regret bound for  the memoryless loss functions $ \{ g_t \} _{t=1}^T $:
\begin{equation*} 
R_T = \sum_{t=1}^T g_t ( x_t ) - \min_{x \in \mathcal{K} }  \sum_{t=1}^T g_t ( x ) = \mathcal{O} \left( T G^2  \eta + \frac{D^2}{\eta} \right).
\end{equation*}
By setting $ \eta = \frac{D}{G\sqrt{T}} $ we get the familiar bound of $ \mathcal{O} \big( GD \sqrt{T} \big) $, which is known to be tight in $G$,$D$ and $T$ against memoryless adversaries.
In addition, if the memoryless loss functions $ \{ g_t \} _{t=1} ^T $ are assumed to be $\sigma$-strongly convex, it is well known that the OGD algorithm attains logarithmic regret bound if $\eta$ is set properly. More specifically,  \cite{HazanAK07} showed that the OGD algorithm generates an online sequence $ \{ x_t \} _{t=1}^T $, for which it holds that:
\begin{equation*} 
R_T = \sum_{t=1}^T g_t ( x_t ) - \min_{x \in \mathcal{K}}  \sum_{t=1}^T g_t ( x ) \leq \sum_{t=1}^T \left\| x_t - x^* \right\| \cdot \left( \frac{1}{\eta_{t+1}} -\frac{1}{\eta_t} - \sigma \right) + G^2 \cdot \sum_{t=1}^T \eta_t  .
\end{equation*}
Setting  $ \eta_t = \frac{1}{\sigma t} $ yields $  R_T \leq \frac{G^2}{\sigma} \left( 1 + \log(T) \right) $.\\

In the framework of OCO with memory, when we allow the loss functions $ \{ f_t \} _{t=1}^T $ to rely on  memory of length $m$, Algorithm \ref{alg:rftlm} with the regularization function $ \mathcal{R}(x) = \| x \| ^2 $ yields the OGD variant for bounded-memory adversaries --- denoted as Algorithm \ref{alg:aogd}. Here, $ \Pi_\mathcal{K}$ refers to the Euclidian projection onto  $ \mathcal{K}$.

\begin{algorithm}[h!]
\caption{OGD with Memory (OGD-M)}
\label{alg:aogd}
\begin{algorithmic}[1]
\STATE Input: learning rate $\eta$, loss functions with memory $\{ f_t \}_{t=1}^T$.
\STATE Choose $x_1, \ldots , x_m \in \mathcal{K}$ arbitrarily.
\FOR {$t=m$ to $T$}
\STATE Play $x_t$ and suffer loss $ f_t (x_{t-m}, \ldots , x_{t}) $.
\STATE Set  $x_{t+1} = \Pi_\mathcal{K} \left( x_t - \eta \nabla \tilde{f}_t (x_t) \right) $.
\ENDFOR
\end{algorithmic}
\end{algorithm}
\noindent We then extend the property of strong convexity to loss functions with memory as follows: we say that $f_t : \mathcal{K}^{m+1} \rightarrow \mathbb{R} $ is $\sigma$-strongly convex loss function with memory if $ \tilde{f}_t (x) = f_t (x, \ldots , x ) $ is $\sigma$-strongly convex in $x$. Thus, for $ \{ f_t \} _{t=1} ^T $ that are $\sigma$-strongly convex loss functions with memory, we can apply Algorithm  \ref{alg:aogd} to get the following result:\\
\begin{corollary} 
Let $\{ f_t \}_{t=1}^T$ be Lipschitz continuous and $\sigma$-strongly convex loss functions with memory (from $\mathcal{K}^{m+1}$ to $[0,1]$), and denote $ G = \sup_{t,x \in \mathcal{K}} \| \nabla  \tilde{f}_t (x)  \|  $. Then, Algorithm \ref{alg:aogd} generates an online sequence $\{ x_t \} _{t=1}^T $, for which the following holds:
\begin{align*}
R_{T,m} & = \sum_{t=m}^T f_t (x_{t-m}, \ldots , x_{t})   -    \min_{x \in \mathcal{K}} \sum_{t=m}^T f_t (x,\ldots,x) \\
& \leq \sum_{t=1}^T \left\| x_t - x^* \right\| \cdot \left( \frac{1}{\eta_{t+1}} -\frac{1}{\eta_t} - \sigma \right) + 2 m^{3/2} G^2 \cdot \sum_{t=1}^T \eta_t  .
\end{align*}
Setting  $ \eta_t = \frac{1}{\sigma t} $ yields $  R_{T,m} \leq \frac{2m^{3/2} G^2}{\sigma} \left( 1 + \log(T) \right) $.
\end{corollary}
The proof simply requires plugging time-dependent learning parameter in the proof of Theorem \ref{rftlm}, and thus omitted here.

\section{Complete Analysis for Section \ref{low}} \label{low_app}
The outline of this section is as follows: we begin by adapting the EWOO algorithm of \cite{HazanAK07} to memoryless convex loss functions (Appendix \ref{adapt}). Then, we present an algorithm for the standard OCO framework that attains low regret and small number of decision switches in expectation (Appendix \ref{analysis}). Finally, we show that these properties together can be reduced to the framework of OCO with memory, yielding a nearly optimal policy regret bound (Appendix \ref{reduce}).

\subsection{Adapting EWOO to Convex Loss Functions} \label{adapt}
Recall the Exponentially Weighted Online Optimization (EWOO) algorithm, presented in \cite{HazanAK07} and designed originally for $\alpha$-exp-concave (memoryless) loss functions $\{ \ell_t \}_{t=1}^T $.
\begin{algorithm}[H]
\caption{Exponentially Weighted Online Optimization (EWOO)}
\label{alg:ewoo}
\begin{algorithmic}[1]
\STATE Input: exp-concavity parameter $\alpha$, exp-concave loss functions $\{ \ell_t \}_{t=1}^T $.
\STATE Initialize $w_1 (x) =1$ for all $x\in\mathcal{K}$, and choose $x_1 \in \mathcal{K}$ arbitrarily.
\FOR {$t=1$ to $T$}
\STATE Play $x_t$ and suffer loss $ \ell_t (x_{t}) $.
\STATE Define weights $w_{t+1} (x)= e^{-\alpha \sum_{\tau=1}^{t} \ell_{\tau} (x)} $.
\STATE Set   $x_{t+1} = \left( \int_{\mathcal{K}} x \cdot w_{t+1} (x) dx \right) \cdot \left( \int_{\mathcal{K}} w_{t+1} (x) dx \right)^{-1}$
\ENDFOR
\end{algorithmic}
\end{algorithm}
\noindent \cite{HazanAK07} prove the following regret bound for Algorithm \ref{alg:ewoo}:
\begin{equation*} \label{regret_ewoo}
R_T = \sum_{t=1}^T \ell_t (x_{t})  - \min_{x \in \mathcal{K}} \sum_{t=1}^T \ell_t (x) \leq \frac{1}{\alpha} \left( 1+ n \log(T+1) \right).
\end{equation*}
Next, we consider the following modification of the EWOO algorithm --- denoted as Algorithm \ref{alg:rewoo}. 
\begin{algorithm}[H]
\caption{}
\label{alg:rewoo}
\begin{algorithmic}[1]
\STATE Input: exp-concavity parameter $\alpha$, exp-concave loss functions $\{ \ell_t \}_{t=1}^T $.
\STATE Initialize $w_1 (x) =1$ for all $x\in\mathcal{K}$, and choose $x_1 \in \mathcal{K}$ arbitrarily.
\FOR {$t=1$ to $T$}
\STATE Play $x_t$ and suffer loss $ \ell_t (x_{t}) $.
\STATE Define weights $w_{t+1} (x)= e^{-\alpha \sum_{\tau=1}^{t} \ell_{\tau} (x)} $.
\STATE Sample   $x_{t+1} $ from the density function $p_t(x) = w_t (x) \cdot \left( \int_{\mathcal{K}} w_{t+1} (x) dx \right)^{-1}$
\ENDFOR
\end{algorithmic}
\end{algorithm}
Basically, $x_t$ is sampled from the density function $ p_t(x) = w_t (x) \cdot \left(  \int_{\mathcal{K}} w_{t} (x) dx  \right)^{-1}$, instead of being computed deterministically. 
The following two lemmas state that applying Algorithm \ref{alg:rewoo} to the loss functions $\{ \hat{g}_t\}_{t=1}^T$ yields regret bound of $\mathcal{O} \big( \sqrt{ T \log(T) } \big)$.
We first bound the regret of Algorithm   \ref{alg:rewoo} when applied to general  $\alpha$-exp-concave loss functions $\{ \ell_t \}_{t=1}^T $ (Lemma \ref{lemma:rewoo}), and then plug in the loss functions $\{ \hat{g}_t\}_{t=1}^T$ (Lemma \ref{convex_ewoo}). \\

\begin{lemma} \label{lemma:rewoo}
Let $\{ \ell_t \}_{t=1}^T$ be $\alpha$-exp-concave loss functions. Then,  Algorithm \ref{alg:rewoo} generates an online sequence $\{ x_t \}_{t=1}^T $, for which the following holds:
\begin{equation*}
\mathbb{E} \left[ R_T \right] = \sum_{t=1}^T \mathbb{E} \left[ \ell_t (x_{t}) \right] - \min_{x} \sum_{t=1}^T \ell_t (x) \leq  \frac{1}{\alpha} \left( 1 + n \log(T+1) \right) +  \frac{\alpha}{2}  \sum_{t=1}^T  \mathbb{E} \left[ \ell_t (x_t)^2  \right] .
\end{equation*}
\end{lemma} 

\begin{proof} The proof goes along the lines of \cite{HazanAK07}; for completeness, we present here the full proof.
Define $ h_t (x) = e^{-\alpha \sum_{\tau =1} ^{t-1} \ell_\tau (x)}$ and notice that
\[
\mathbb{E} \left[ h_t (x_t) \right] = \int_\mathcal{K} h_t (x)  p_t (x) dx = \frac{ \int_\mathcal{K} \left( \prod_{\tau=1}^t h_\tau (x) \right) dx  }{ \int_\mathcal{K} \left( \prod_{\tau=1}^{t-1} h_\tau (x) \right) dx} .
\]
Then, by telescopic product we have 
\begin{equation} \label{telescopic}
\prod_{t=1}^{T} \mathbb{E} \left[ h_t (x_t) \right]  =  \frac{ \int_\mathcal{K} \left( \prod_{t=1}^T h_t (x) \right) dx  }{ \int_\mathcal{K} 1 dx} =  \frac{ \int_\mathcal{K} \left( \prod_{t=1}^T h_t (x) \right) dx  }{ vol \left( \mathcal{K} \right) } ,
\end{equation}
where we used the fact that $w_1 (x)=1$ for all $x\in\mathcal{K}$. Denote $x^* = \arg \min _{x\in\mathcal{K}} \sum_{t=1}^T \ell_t (x) $, then it exists that $x^* = \arg \max _{x\in\mathcal{K}} \prod_{t=1}^{T}  h_t (x)  $. Define nearby points $\mathcal{S} \subset \mathcal{K}$ by
\[
\mathcal{S} = \left\{ x \in \mathcal{K} \mid x = \frac{T}{T+1} x^* + \frac{1}{T} y \ \ , \ \ y \in \mathcal{K}\right\} . 
\]
By concavity and non-negativity of $h_t$ it holds that $h_t (x) \geq \frac{T}{T+1}  h_t (x^*)$
for every $x \in \mathcal{S}$, and thus
\[
\prod_{t=1}^{T}  h_t (x)   \geq \left( \frac{T}{T+1} \right)^T  \prod_{t=1}^{T} h_t (x^*) \geq  e^{-1} \prod_{t=1}^{T} h_t (x^*)  .
\]
By substituting the above in Equation \eqref{telescopic} and using the fact that $\mathcal{S}$ is a rescaling of $\mathcal{K}$ by factor of $\frac{1}{T+1}$ in $n$ dimensions, we have that
\begin{align*}
 \prod_{t=1}^{T} \mathbb{E} \left[ h_t (x_t) \right] &  =  \frac{ \int_\mathcal{K} \left( \prod_{t=1}^T h_t (x) \right) dx  }{ vol \left( \mathcal{K} \right) }  
 \geq  \frac{ \int_\mathcal{S} \left( \prod_{t=1}^T h_t (x) \right) dx  }{ vol \left( \mathcal{K} \right) } \\ &  \geq  \frac{ \int_\mathcal{S} \left( e^{-1} \prod_{t=1}^T h_t (x^*) \right) dx  }{ vol \left( \mathcal{K} \right) }   = \frac{  vol ( \mathcal{S})  }{ vol\left(\mathcal{K} \right) }   e^{-1}  \prod_{t=1}^T h_t (x^*) \\& = \frac{ e^{-1}  }{ (T+1)^n }  \prod_{t=1}^T h_t (x^*).
\end{align*}
Now, by taking logarithm on both sides we get that
\begin{equation*}
 \sum_{t=1}^T \log \left( \mathbb{E} \left[ h_t(x_t) \right] \right) -  \sum_{t=1}^T \log \left( h_t (x^*) \right) \geq -1 - n \log(T+1) ,
\end{equation*}
or equivalently
\begin{equation} \label{log}
 \sum_{t=1}^T \log \left( \mathbb{E} \left[ e^{-\alpha \ell_t  (x_t)} \right] \right) + \alpha  \sum_{t=1}^T \ell_t  (x^*) \geq -1 - n \log(T+1) .
\end{equation}
Next, we use the facts that $ e^{-x} \leq 1 - x +\frac{x^2}{2}$ for $ 0\leq x \leq1$ and $\log(1-x) \leq -x$ for  $x <1$, to derive the following inequality:
\begin{align*}
 \log  \left( \mathbb{E} \left[ e^{-\alpha  \ell_t (x_t)} \right] \right) & \leq   \log \left( \mathbb{E} \left[ 1-\alpha  \ell_t (x_t) + \frac{\alpha^2}{2}  \ell_t (x_t)^2  \right] \right)  \\
&  =   \log \left( 1-\alpha \mathbb{E} \left[ \ell_t  (x_t) \right] +  \frac{\alpha^2}{2} \mathbb{E} \left[  \ell_t (x_t)^2  \right]  \right) \\
& \leq -\alpha \mathbb{E} \left[  \ell_t  (x_t) \right] +  \frac{\alpha^2}{2} \mathbb{E} \left[  \ell_t  (x_t)^2  \right] 
\end{align*}
By substituting the above in Equation \eqref{log} and rearanging we get that
\begin{equation*} 
 \sum_{t=1}^T \mathbb{E} \left[ \ell_t  (x_t) \right]   -  \sum_{t=1}^T  \ell_t  (x^*) \leq \frac{1}{\alpha} \left( 1 + n \log(T+1) \right) +  \frac{\alpha}{2}  \sum_{t=1}^T  \mathbb{E} \left[  \ell_t  (x_t)^2  \right] ,
\end{equation*}
as stated in the lemma.
\end{proof}

\noindent Plugging in the loss functions $\{ \hat{g}_t\}_{t=1}^T$ into the previous lemma yields the following result:\\

\begin{lemma} \label{convex_ewoo}
Let $\{ g_t \}_{t=1}^T$ be convex functions from $\mathcal{K} $ to $ [0,1] $, such that  $D =  \sup_{x,y \in \mathcal{K}} \| x-y\|  $ and $ G =  \sup_{x,t} \| \nabla  g_t (x)  \| , $ and define $\hat{g}_t(x) = g_t(x) + \frac{\eta}{2} \| x\|^2$ for some $\eta \leq \frac{G}{D}$.  Then, Applying Algorithm \ref{alg:rewoo} to the loss functions $ \{ \hat{g}_t \}_{t=1}^T $ generates an online sequence $\{x_t\}_{t=1}^T$, for which the following holds:
\begin{equation*}
\mathbb{E} \left[ R_T \right] = \sum_{t=1}^T \mathbb{E} \left[ {g}_t (x_{t})  \right]  - \min_{x} \sum_{t=1}^T {g}_t (x)  \leq \frac{4G^2}{\eta}  \left( 1 + n \log(T+1) \right) + \frac{T \eta }{2}  \left( \frac{ \left( 1+\eta D^2  \right)^2 }{4G^2} +  D^2   \right).
\end{equation*}
Setting $ \eta = \frac{2G}{D} \sqrt{ \frac{1+ \log(T+1)}{T} } $ yields $\mathbb{E} \left[ R_T \right]  \leq 8n \cdot  \max  \left\{ GD , \frac{1}{GD} \right\} \cdot \sqrt{ T (1+\log(T+1)) }$.
\end{lemma} 

\begin{proof}
Recall that the loss functions $\{ \hat{g}_t \}_{t=1}^T$ are $\frac{\eta}{4G^2}$-exp-concave for $\eta \leq \frac{G}{D}$. Thus, applying Algorithm \ref{alg:rewoo} to the loss functions $\{ \hat{g}_t \}_{t=1}^T$ yields the following result (using Lemma \ref{lemma:rewoo}):
\begin{equation*}
\sum_{t=1}^T \mathbb{E} \left[ \hat{g}_t (x_{t})  \right]  - \min_{x} \sum_{t=1}^T \hat{g}_t (x)  \leq \frac{4G^2}{\eta}  \left( 1 + n \log(T+1) \right) + \frac{\eta}{8G^2}  \sum_{t=1}^T  \mathbb{E} \left[ \hat{g}_t (x_t)^2  \right] .
\end{equation*}
By substituting $\hat{g}_t$ from the definition and using the fact that $ \hat{g}_t (x) \in [0,1+\eta D^2]$ for all $t$ and  $ x \in \mathcal{K}$, we have that
\begin{align*}
 \sum_{t=1}^T \mathbb{E} & \left[ {g}_t (x_{t})  \right]  - \min_{x} \sum_{t=1}^T {g}_t (x)  \\
&  \leq \frac{4G^2}{\eta}  \left( 1 + n \log(T+1) \right) + \frac{\eta}{2}  \sum_{t=1}^T \left( \frac{ \left( 1+\eta D^2  \right)^2 }{4G^2} +  \| x^* \|^2 -\| x_t \|^2   \right).
\end{align*}
The lemma is obtained by observing that $ \| x^* \|^2 -\| x_t \|^2 \leq D^2 $.
\end{proof}

 \subsection{Algorithm and Analysis} \label{analysis}
 

We turn now to restate and prove our main theorem:\\

\noindent \textbf{Theorem 4.1.}
\textit{Let $\{ g_t \}_{t=1}^T$ be convex functions from $\mathcal{K} $ to $ [0,1] $, such that 
$D =  \sup_{x,y \in \mathcal{K}} \| x-y\| $ and $ G =  \sup_{x,t} \| \nabla  g_t (x)  \| $,
 and define $\hat{g}_t(x) = g_t(x) + \frac{\eta}{2} \| x\|^2$ for some $\eta \leq \frac{G}{D}$.  Then,  Algorithm \ref{alg:rewoo_low}  generates an online sequence $\{ x_t\}_{t=1}^T$, for which it holds that}
$$ \mathbb{E} \left[ R_T \right] = \sum_{t=1}^T \mathbb{E} \left[ g_t (x_{t})  \right]  - \min_{x \in \mathcal{K}} \sum_{t=1}^T g_t (x)  \leq \frac{4G^2}{\eta}  \left( 1 + n \log(T+1) \right) + \frac{T \eta }{2}  \left( \frac{ \left( 1+\eta D^2  \right)^2 }{4G^2} +  D^2   \right) ,$$
\textit{and in addition}
$$ \mathbb{E} \left[ S \right]  =\mathbb{E} \left[  \sum_{t=1}^T 1_{ \{ x_{t+1} \neq x_{t} \} }  \right]  \leq \frac{T \eta}{4G^2} + \frac{T D^2 \eta^2}{8G^2}  ,  $$

\textit{Setting $\eta = \frac{2G}{D} \sqrt{ \frac{1+ \log(T+1)}{T} } $ yields $\mathbb{E} \left[ R_T \right]  = \mathcal{O} \big( \sqrt{ T \log(T) } \big)$, and $ \mathbb{E} \left[ S \right] = \mathcal{O} \big( \sqrt{ T \log(T) } \big)$.
 }\\

%

\begin{proof}
The proof follows immediately by observing that: (1)  Algorithm \ref{alg:rewoo_low}  generates the decisions from the same distribution with respect Algorithm \ref{alg:rewoo} (stated formally in Lemma \ref{dist} below), and thus attains the same expected regret bound; and (2)  Algorithm \ref{alg:rewoo_low} has an expected low switches guarantee (also stated below in Lemma \ref{switches}). 
\end{proof}
We shall continue to prove the lemmas.\\
\begin{lemma} \label{dist}
Let $\{ g_t \}_{t=1}^T$ be convex functions from $\mathcal{K} $ to $ [0,1] $, such that  $D =  \sup_{x,y \in \mathcal{K}} \| x-y\|$ and $ G =  \sup_{x,t} \| \nabla  g_t (x)  \| $, and define $\hat{g}_t(x) = g_t(x) + \frac{\eta}{2} \| x\|^2$ for some $\eta \leq \frac{G}{D}$. Denote by $\{ y_t \}_{t=1}^T$ and $\{ x_t \}_{t=1}^T$ the online sequences generated by applying Algorithm  \ref{alg:rewoo_low} and Algorithm  \ref{alg:rewoo} to the loss functions  $ \{ {g}_t \}_{t=1}^T $ and $ \{ \hat{g}_t \}_{t=1}^T $, respectively. Then, it holds that $y_t$ and $x_t$ are sampled from the same distribution for all $t$.\\
\end{lemma}

\begin{proof}
Let $q_t(\cdot)$ and $p_t(\cdot)$ be the density functions of $y_t$ and $x_t$, respectively, and  $W_t = \int_{\mathcal{K}} w_{t} (x) dx$. The proof is by induction: for $t=1$ we have from the definition that $p_1(x)=q_1(x)$ for all $x \in \mathcal{K}$. Now, let us assume that $p_{t-1}(x)=q_{t-1}(x)$ for all $x \in \mathcal{K}$, and prove for $t$.
 Notice that the weights update for both algorithms is the same and is independent of the decisions actually played by the player.
  Thus, by applying the law of total probability we have that 
\begin{align*}
 q_t (x) & = p_{t-1} (x) \cdot \frac{w_t(x)}{w_{t-1}(x)} + p_t(x) \cdot \int_{\mathcal{K}} p_{t-1} (y) \left( 1- \frac{w_t(y)}{w_{t-1}(y)} \right)dy \nonumber\\
&  =  \frac{w_{t-1}(x)}{W_{t-1}} \cdot \frac{w_t(x)}{w_{t-1}(x)} + \frac{w_{t}(x)}{W_{t}} \cdot \int_{\mathcal{K}}  \frac{w_{t-1}(y)}{W_{t-1}}  \left( \frac{w_{t-1}(y)-w_t(y)}{w_{t-1}(y)} \right)dy \nonumber\\
&  =  \frac{w_{t}(x)}{W_{t-1}}  + \frac{w_{t}(x)}{W_{t}} \cdot \int_{\mathcal{K}}  \frac{w_{t-1}(y)-w_t(y)}{W_{t-1}} dy  \nonumber\\
& =  \frac{w_{t}(x)}{W_{t-1}}  + \frac{w_{t}(x)}{W_{t}} \cdot  \frac{W_{t-1} - W_t}{W_{t-1}}  \nonumber\\
& =  \frac{w_{t}(x) \cdot W_{t}+w_{t}(x) \cdot W_{t-1} - w_{t}(x) \cdot W_{t}}{W_{t-1} \cdot W_{t}} \nonumber\\
&  = \frac{w_{t}(x) \cdot W_{t-1} }{W_{t-1} \cdot W_{t}} = \frac{w_{t}(x) }{W_{t}} = p_t(x).
\end{align*}
The above holds for all $x \in \mathcal{K}$, and thus the lemma is obtained.
\end{proof}

\begin{lemma} \label{switches}
Let $\{ g_t \}_{t=1}^T$ be convex functions from $\mathcal{K} $ to $ [0,1] $, such that  $D =  \sup_{x,y \in \mathcal{K}} \| x-y\|  $ and $ G =  \sup_{x,t} \| \nabla  g_t (x)  \| $. Then, applying Algorithm \ref{alg:rewoo_low}  to the loss functions $ \{ {g}_t \}_{t=1}^T $ generates an online sequence $ \{ x_t \}_{t=1}^T $, for which the it holds that
\begin{equation*}
\mathbb{E}  \left[ S \right] =  \sum_{t=1}^T \mathbb{E} \left[ 1_{ \{ x_{t+1} \neq x_{t} \} }  \right]   \leq  \frac{T \eta}{4G^2} + \frac{T D^2 \eta^2}{8G^2}  ,
\end{equation*}
where $S$ denotes the number of decision switches in the sequence $ \{ x_t \} _{t=1}^T$. 

\noindent Setting $\eta = \frac{2G}{D} \sqrt{ \frac{1+ \log(T+1)}{T} } $ yields $ \mathbb{E}  \left[ S \right]    \leq  1 + \log(T+1) + \frac{1}{GD} \sqrt{T \left( 1+ \log(T+1) \right)} $.
\end{lemma}

\begin{proof}
From Algorithm \ref{alg:rewoo_low} it follows that
\begin{equation*}
\mathbb{E} \left[ 1_{ \{ x_{t+1} \neq x_{t} \} }  \right]   = P \left(  x_{t+1} \neq x_{t} \right) \leq 1-\frac{w_{t+1}(x_t)}{w_t(x_t)} = 1-e^{-\frac{\eta}{4G^2} \hat{g}_t (x_t)},
\end{equation*}
Using the inequality $ 1 - e^{-x} \leq x $ for all $x$, and substituting $\hat{g}_t$ from the definition yields
\begin{equation*}
1-e^{-\frac{\eta}{4G^2} \hat{g}_t (x_t)} \leq  \frac{\eta}{4G^2} {g}_t (x_t) + \frac{\eta^2}{8G^2} \| x_t \|^2 .
\end{equation*}
Next, by summing the above for all $t$ we have that
\begin{equation*}
 \sum_{t=1}^T \mathbb{E} \left[ 1_{ \{ x_{t+1} \neq x_{t} \} }  \right]   \leq \frac{\eta}{4G^2} \sum_{t=1}^T  {g}_t (x_t) + \frac{\eta^2}{8G^2} \sum_{t=1}^T \| x_t \|^2 .
\end{equation*}
Finally, since $ \| x \|^2 \leq D^2 $ for all $x\in \mathcal{K}$ and $g_t(x) \in [0,1]$ for all $x\in \mathcal{K}$ and $t \in \{ 1, \ldots,T\}$, setting $\eta = \frac{2G}{D} \sqrt{ \frac{1+ \log(T+1)}{T} } $ gives the stated result.
\end{proof}

\subsection{Reduction to the Framework of OCO with Memory} \label{reduce}
Up to this point, we presented an algorithm that attains  $\mathcal{O} \big( \sqrt{ T \log(T) } \big)$-regret along with expected $\mathcal{O} \big( \sqrt{ T \log(T) } \big)$ decision switches for generally convex loss functions $\{ g_t \} _{t=1}^T$. The next lemma states that these two properties imply learning against bounded-memory adversaries. \\

\begin{lemma} \label{reduction}
Let $\{ f_t \}_{t=1}^T$ be loss functions with memory from $\mathcal{K}^{m+1} $ to $ [0,1] $,  define $ \tilde{f}_t (x) = f_t (x,\ldots,x)$, and denote $ D =  \sup_{x,y \in \mathcal{K}} \| x-y\| $ and $ G =  \sup_{x ,t } \| \nabla   \tilde{f}_t (x)  \| $.
Then, applying Algorithm \ref{alg:rewoo_low} to the loss functions  $\{ \tilde{f}_t \}_{t=1}^T$  yields an online sequence $\{ x_t\}_{t=1}^T$, for which it holds that:
\begin{align*}
 \mathbb{E} \left[ R_{T,m} \right] & = \sum_{t=1}^T \mathbb{E} \left[ f_t (x_{t-m}, \ldots , x_{t})   \right]  - \min_{x \in \mathcal{K}} \sum_{t=1}^T f_t (x,\ldots,x) \nonumber\\
&   \leq \frac{4G^2}{\eta}  \left( 1 + n \log(T+1) \right) + \frac{T \eta }{2}  \left( \frac{ \left( 1+\eta D^2  \right)^2 }{4G^2} +  D^2   \right)  + \frac{T m \eta}{4G^2} + \frac{T D^2 m \eta^2}{8G^2}  .
\end{align*}
Setting $\eta = \frac{2G}{D} \sqrt{ \frac{1+ \log(T+1)}{mT} } $ yields $\mathbb{E} \left[ R_{T,m} \right]  \leq 8n \cdot  \max  \left\{ GD , \frac{1}{GD} \right\} \cdot \sqrt{ m T (1+\log(T+1)) }$.
\end{lemma}

\begin{proof}
From Theorem \ref{main2}, we know that applying Algorithm  \ref{alg:rewoo_low} to the loss functions  $\{ \tilde{f}_t \}_{t=1}^T$ yields:
$$  \sum_{t=1}^T \mathbb{E} [ \tilde{f}_t (x_{t})  ]  - \min_{x \in \mathcal{K}} \sum_{t=1}^T \tilde{f}_t (x)  \leq \frac{4G^2}{\eta}  \left( 1 + n \log(T+1) \right) + \frac{T \eta }{2}  \left( \frac{ \left( 1+\eta D^2  \right)^2 }{4G^2} +  D^2   \right) ,$$
or equivalently:
\begin{align} \label{eq1}
 \sum_{t=1}^T   \mathbb{E} & \left[  f_t (x_{t}, \ldots , x_{t})   \right] - \min_{x  \in \mathcal{K}}  \sum_{t=1}^T  f_t (x,\ldots,x)  \nonumber\\
 & \leq  \frac{4G^2}{\eta}  \left( 1 + n \log(T+1) \right)  + \frac{T \eta }{2}  \left( \frac{ \left( 1+\eta D^2  \right)^2 }{4G^2} +  D^2   \right).
\end{align}
Now, notice that if a decision switch did not occur between time points $ (t-m) $ and $ t $, it trivially holds that 
$
f_t (x_{t-m}, \ldots , x_{t}) = f_t (x_{t}, \ldots , x_{t}) .
$
Otherwise, if a decision switch did occur between these time points, we can bound
$
| f_t (x_{t-m}, \ldots , x_{t}) = f_t (x_{t}, \ldots , x_{t}) | \leq 1.
$
Thus, it follows that
$$
 \sum_{t=m}^T \left| f_t (x_{t-m}, \ldots , x_{t})   - 
f_t (x_{t}, \ldots , x_{t})  \right| \leq m \cdot S ,
$$
where again, $S$ denotes the number of decision switches in the sequence $\{x_t\}_{t=1}^T$. From Lemma \ref{switches} we have that
$ \mathbb{E}  \left[ S \right]  \leq \frac{T \eta}{4G^2} + \frac{T D^2 \eta^2}{8G^2}  $, and it follows that
\begin{align*}
  \left|  \sum_{t=m}^T \mathbb{E} \left[ f_t (x_{t-m}, \ldots , x_{t}) \right]  - \sum_{t=m}^T \mathbb{E} \left[ f_t (x_{t}, \ldots , x_{t}) \right]  \right|  & \leq \sum_{t=1}^T  \left| \mathbb{E} \left[ f_t (x_{t-m}, \ldots , x_{t})   - f_t (x_t,\ldots,x_t)  \right] \right| \nonumber\\
&   \leq \sum_{t=1}^T   \mathbb{E} \left[ \left| f_t (x_{t-m}, \ldots , x_{t})    -  f_t (x_t,\ldots,x_t) \right|  \right]  \nonumber\\
& \leq m \cdot \mathbb{E} \left[ S \right] \leq \frac{T m \eta}{4G^2} + \frac{T D^2 m \eta^2}{8G^2} .
\end{align*}
Plugging the above in Equation \eqref{eq1} yields the result stated in the lemma.
\end{proof}

\section{Efficient Implementation of Algorithm \ref{alg:rewoo_low}} \label{imp}
The original EWOO algorithm (Algorithm \ref{alg:ewoo}) of \cite{HazanAK07} is not efficient, since it generates $x_t$ as the expectation with respect to the distribution $p_t$ in every iteration. \textit{Hazan et al.} solve this issue by referring to the works of \cite{DBLP:conf/focs/LovaszV03a}, that offer a sampling method from logconcave distributions. These  techniques enable the sampling of $m$ points from the distribution $p_t$ in time of $ \tilde{\mathcal{O}} ( n^4 + m n^3 ) $.  Since an accuracy of $T^{-1}$ to the expectation is necessary for maintaining logarithmic regret, $m$ must be on the order of $T^2$. Thus, generating a single decision $x_t$ via a slightly modified EWOO algorithm requires running time of $ \tilde{\mathcal{O}} ( n^4 + T^2 n^3 ) $, which results in a total running time of $ \tilde{\mathcal{O}} ( T n^4 + T^3 n^3 ) $.

The implementation of the proposed algorithm (Algorithm \ref{alg:rewoo_low}) can rely on the same techniques as algorithm EWOO, yet can be carried out more efficiently in various ways. First, our algorithm requires only $ \tilde{\mathcal{O}} ({T}^{1/2})$ samples (in compare to $T$ samples that EWOO requires), due to its low switches guarantee. Second, each of these samples requires time of $ \tilde{\mathcal{O}} ( n^4 ) $ using the techniques of  \cite{DBLP:conf/focs/LovaszV03a}, because $x_t$ need not be generated as the expectation of $p_t$, but rather only be sampled from this distribution.  Therefore, an efficient implementation of our algorithm can be carried out in a total running time of $ \tilde{\mathcal{O}} ( T^{1/2} n^4 ) $.

Another efficient implementation of Algorithm \ref{alg:rewoo_low} relies on the work of \cite{DBLP:conf/nips/NarayananR10}, in which techniques of random walks are utilized for regret minimization. Basically, these techniques are applicable in our setting for two reasons: (1) two successive distributions over the decision set, $p_t$ and $p_{t+1}$, are relatively close; and (2) each distribution $p_t$ can be approximated quite well using a  Gaussian distribution. This allows sampling $x_{t+1} $ via a random walk technique that requires only one step, due to the fact that $x_t$ can be used as its warm start. This results in a same running time guarantee for our algorithm, as stated before for the techniques of  \cite{DBLP:conf/focs/LovaszV03a}.

\end{document}